\documentclass[letterpaper, 10 pt, conference]{ieeeconf}
\IEEEoverridecommandlockouts                              

\usepackage{graphicx}
\usepackage{caption}
\usepackage{amsmath,amssymb,enumerate}

\usepackage{amsthm}
\usepackage{mathtools}
\usepackage{breqn}
\usepackage{algorithm, algorithmicx, algpseudocode}

\usepackage{blindtext}
\usepackage{gensymb}
\usepackage{xparse}
\usepackage{lipsum}
\usepackage{mathrsfs}
\usepackage[mathscr]{euscript}
\usepackage{times}
\usepackage{cite} 
\usepackage{multicol}
\usepackage[caption=false,font=footnotesize]{subfig}
\usepackage{amsfonts}
\usepackage[utf8]{inputenc}
\usepackage[T1]{fontenc}
\usepackage{textcomp}
\usepackage{amsfonts}
\usepackage{soul}
\usepackage{multirow}
\usepackage{booktabs}
\usepackage[usenames,dvipsnames,svgnames,table, xcdraw]{xcolor}

\usepackage[colorlinks=true,pdfpagemode=UseNone,citecolor=black,linkcolor=black,urlcolor=BrickRed]{hyperref}

\usepackage{comment}

\newtheorem{problem}{Problem}
\newtheorem{theorem}{Theorem}
\newtheorem{corollary}[theorem]{Corollary}

\newtheorem{remark}{Remark}

\DeclareMathOperator*{\argmax}{arg\,max}

\renewcommand{\thefigure}{\arabic{figure}}

\newcommand{\m}{\mathop{\mathrm{m}}}

\newcommand{\Hcal}{\mathcal{H}}
\newcommand{\Ical}{\mathcal{I}}

\title{\LARGE \bf A New Framework for Registration of Semantic Point Clouds from Stereo and RGB-D Cameras}

\author{Ray Zhang, Tzu-Yuan Lin, Chien Erh Lin, Steven A. Parkison, William Clark, Jessy W. Grizzle, \\
Ryan M. Eustice and Maani Ghaffari

\thanks{*Toyota Research Institute (TRI) provided funds to support this work.}%
\thanks{R.~Zhang, T.Y.~Lin, C.E.~Lin, J.~Grizzle, R.~Eustice, and M.~Ghaffari are with the University of Michigan, Ann Arbor, MI 48109, USA. {\tt\small\{rzh,tzuyuan,chienerh,grizzle,eustice,maanigj\} @umich.edu}}%
\thanks{S.~Parkison is with TRI. {\tt\small steven.parkison@tri.global}}%
\thanks{W.~Clark is with the department of Mathematics, Cornell University, Ithaca, NY. {\tt\small wac76@cornell.edu}}
}

\begin{document}

\maketitle
\thispagestyle{empty}
\pagestyle{empty}

\begin{abstract}
This paper reports on a novel nonparametric rigid point cloud registration framework that jointly integrates geometric and semantic measurements such as color or semantic labels into the alignment process and does not require explicit data association. The point clouds are represented as nonparametric functions in a reproducible kernel Hilbert space. The alignment problem is formulated as maximizing the inner product between two functions, essentially a sum of weighted kernels, each of which exploits the local geometric and semantic features. As a result of the continuous models, analytical gradients can be computed, and a local solution can be obtained by optimization over the rigid body transformation group. Besides, we present a new point cloud alignment metric that is intrinsic to the proposed framework and takes into account geometric and semantic information. The evaluations using publicly available stereo and RGB-D datasets show that the proposed method outperforms state-of-the-art outdoor and indoor frame-to-frame registration methods. An open-source GPU implementation is also provided.
\end{abstract}

\section{INTRODUCTION}

\begin{figure}[t]
  \centering 
  \includegraphics[width=0.92\columnwidth,trim={0.0cm 0.75cm 0.0cm 2cm},clip]{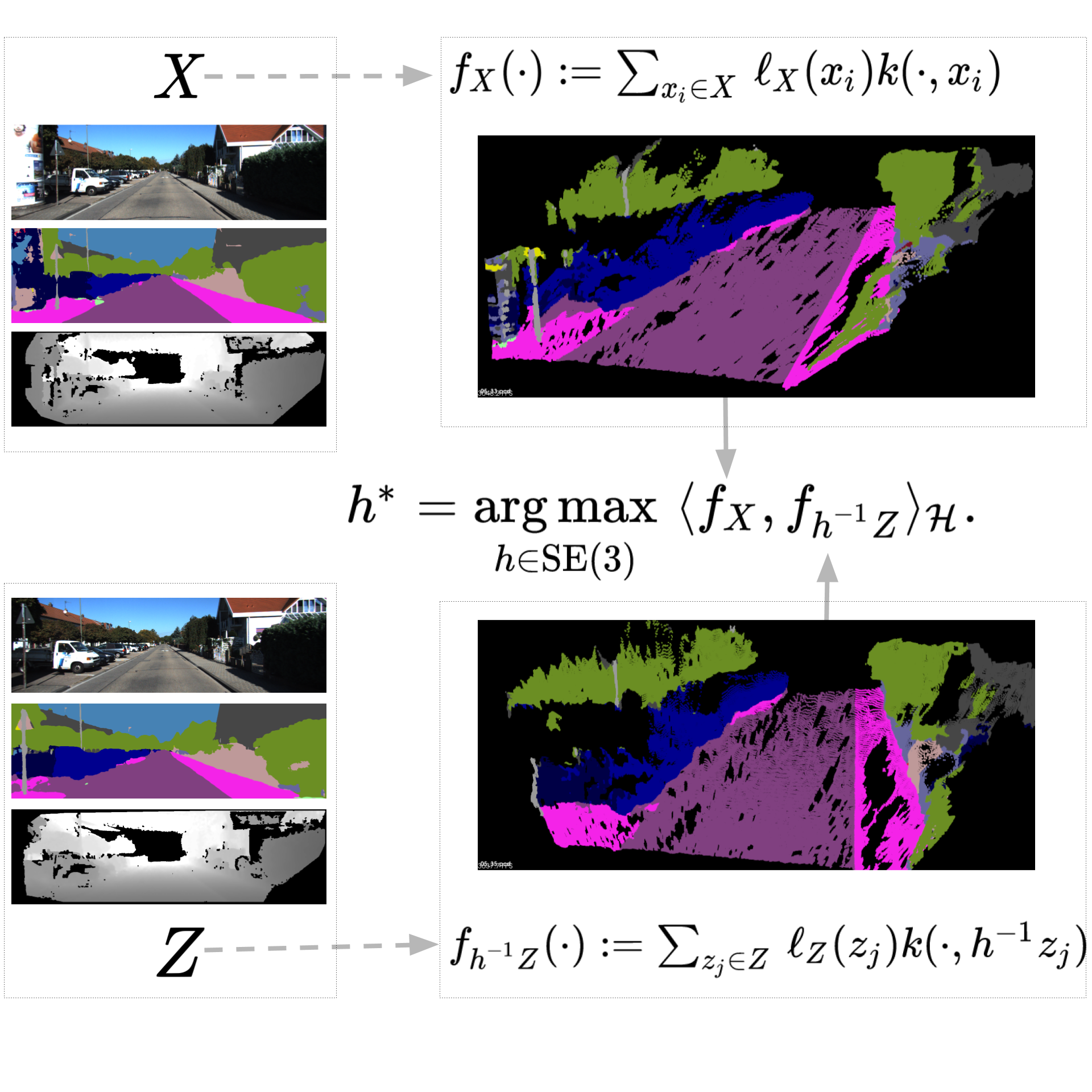}\vspace{-0.5cm}
  \caption{Point clouds $X$ and $Z$ are represented by two continuous functions $f_X, f_Z$ in a reproducing kernel Hilbert Space. Each point $x_i$ has its own semantic labels, $\ell_X(x_i)$, encoded in the corresponding function representation via a tensor product representation. The registration is formulated as maximizing the inner product between two point cloud functions.}
 \label{fig:overview_cvo}
\end{figure}

Point cloud registration estimates the relative transformation between two noisy point clouds
\cite{besl1992icp,chen1992surfacenormalicp,segal2009gicp,pomerleau15review1,cheng2018registration}. Point clouds obtained by RGB-D cameras, stereo cameras, and LIDARs contain rich color and intensity measurements besides the geometric information. The extra non-geometric information can improve the registration performance~\cite{servos2014mcicp,parkison2018semanticicp,parkison2019rkhsicp}. Deep learning can provide semantic attributes of the scene as measurements~\cite{long2015fcn,chen2017deeplab,zhu2019improving}. As illustrated in Fig.~\ref{fig:overview_cvo}, this work focuses on the construction of a novel integrated framework to jointly process raw geometric and non-geometric information for point cloud registration.

\begin{figure}[t]
  \centering
  \vspace{0.2cm}
 \subfloat{\includegraphics[width=0.99\columnwidth,clip]{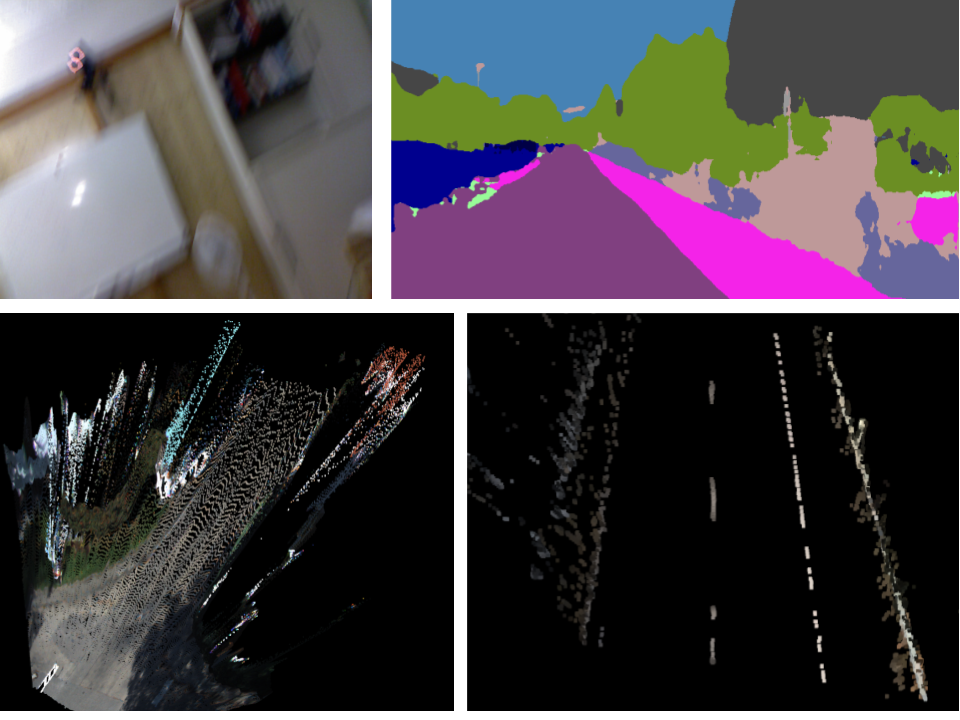}}
  
  \caption{Challenging scenes for stereo and RGB-D point cloud registration, including blurry image sources (from TUM RGB-D~\cite{sturm12iros}), noisy semantic sources (from KITTI~\cite{kitti2012} and Nvidia~\cite{zhu2019improving}), noisy depth estimations (from KITTI), and highly repetitive patterns (from KITTI).}
 \label{fig:difficult_scenes}
\end{figure}

Real-world applications such as SLAM\cite{whelan2016elasticfusion} and 3D reconstruction\cite{zollhofer2018state3dreconstruct} include noisy measurements, symmetries or dynamics objects, occlusion, and blurry observations. Examples are shown in Fig.~\ref{fig:difficult_scenes}. These cases make the data association process challenging. Existing Iterative Closest Point (ICP)-based work~\cite{besl1992icp,chen1992surfacenormalicp,segal2009gicp} approach this problem by adding appearance/semantic features~\cite{park2017colored,servos2014mcicp,parkison2018semanticicp},  adding local or deep geometric features~\cite{segal2009gicp,FCGF2019}, and introducing weighted many-to-many correspondences~\cite{gold1998softassign,granger02emicp}. 

Gaussian Mixture Model (GMM) based registrations \cite{biber2003ndt2d,magnusson2007ndt3d, jian2011gmm} model data correspondences and point clouds as probabilistic densities. Instead of point pairs, GMM-based methods work on point clusters, and the associations are a part of the model parameters. The relative rigid body transformation is then estimated by fitting the second point cloud measurements into the first point cloud's distributions~\cite{magnusson2007ndt3d,chui2000mpm, horaud2010ecmpr,eckart2015mlmd,eckart2013remseg,eckart2018hgmr}, or by minimizing a distance measure between the two distribution and inferring the weight parameters~\cite{tsin2004correlation,wang06cdfjs,jian05gmmreg}.

This paper  presents a nonparametric registration framework that jointly integrates geometric and semantic measurements and does not require explicit data association. 
Unlike existing methods that rely on geometric residuals with regularizers to include appearance information~\cite{park2017colored,parkison19rvmicp}, the proposed framework formulates the problem using a single objective function, and is solved by the gradient ascent on Riemannian manifolds, similar to the work of~\cite{MGhaffari-RSS-19,clarkmaani20}. In particular, this work has the following contributions. 
\begin{enumerate}
    \item A novel framework for semantic point cloud registration that generalizes geometric, color, and semantic-assisted methods to a nonparametric continuous model via a hierarchical distributed representation of features.
    \item A new point cloud alignment indicator that is intrinsic to the proposed framework and takes into account geometric and semantic information.
    \item An open source GPU implementation available at~\cite{unified-cvo}:\\ \href{https://github.com/UMich-CURLY/unified_cvo}{https://github.com/UMich-CURLY/unified_cvo}
    \item Extensive evaluations using publicly available datasets for outdoor stereo and indoor RGB-D datasets.
\end{enumerate}

\section{Related Work}
\label{sec:related_work}

To improve the quality of one-to-one correspondences (\emph{hard} assignment), the work of \cite{chetverikov2005robustalignment} assumes that only a portion of points can be paired thus only considers first few smallest residuals. Many-to-many correspondences (\emph{soft} assignment) are introduced as the weights of the residuals, controlling the "blurriness" of point matches. The weights can come from mutual information\cite{rangarajan1999rigid} or from Gaussian weights \cite{gold1998softassign}.  EM-ICP\cite{granger02emicp} treats the correspondences as hidden variables, and use Expectation Maximization (EM) \cite{bishop2006pattern} to infer both the matches and then the transformations. 

Point-to-plane\cite{chen1992surfacenormalicp}, plane-to-plane\cite{mitra04_surface_icp}, and Generalized-ICP\cite{segal2009gicp} build local geometric structures to the loss formulation. The work of \cite{sharp2002invariant_icp} combines multiple Euclidean invariant features. 
The work of~\cite{Sehgal2010sift} works on an IR camera and uses extra SIFT features from depth images to help keypoint correspondences. The work of~\cite{servos2014mcicp} uses color/intensity for both association and registration. Color ICP~\cite{park2017colored} defines a sum of reprojected photometric and depth loss on dense RGB-D point clouds. GICP-RKHS~\cite{parkison19rvmicp} also appends an additional regularizer to the GICP's loss for point intensity via the Relevance Vector Machine~\cite{bishop2006pattern}. Semantic-ICP~\cite{parkison2018semanticicp} treats points' semantic labels and associations as additional hidden variables as a part of the EM-ICP framework. In our formulation, function representation combines both geometric and non-geometric information into a unified formulation, and it affects both the association and the optimization steps. 

\subsection{Mixture of Gaussian-based Registration Frameworks}
Probabilistic registration frameworks represent point clouds as discrete\cite{biber2003ndt,magnusson2007ndt3d} or continuous probability densities \cite{jian2011gmm,campbell2015svr,chui2000mpm,eckart2018hgmr,evangelidis2017joint,horaud2010ecmpr}. Compared to this work, GMM based methods also use a double sum of Gaussian kernels, combined with the soft data association, but they come from a different theoretical background than the Reproducing Kernel Hilbert Space (RKHS)~\cite{berlinet04rkhsbook}.

Normal Distribution Transform (NDT) defines a discrete collection of  bivariate Gaussian distributions to capture local surface structures~\cite{biber2004probabilistic,magnusson2007ndt3d}.  Discretization brings automatic soft data association without the need of inferring GMM weights. An effective discretization strategy requires  suitable voxel sizes and efficient voxel deployment, for instance  a forest of octrees\cite{magnusson2007ndt3d}, distance based voxel sizing\cite{miao16}, hierarchical voxel tree deployment \cite{ulacs2013multilayerndt}, and cell clustering\cite{das2014scanndt}.
Comparing to NDT, the proposed method is also data association free, but it is further a continuous representation, thus avoids the above concerns caused by discretization. 

Some continuous GMM-based methods minimize the distance between two distributions. Effective distance measures include Jensen-Shannon divergence\cite{wang06cdfjs} and the
$l_2$ distance of the two dense \cite{jian05gmmreg} or sparse\cite{campbell2015svr} GMMs.  Kernel Correlation (KC) \cite{tsin2004correlation} maximizes the correlation between two point clouds using M-estimators, in particular a sum of Gaussian kernels. It has an identical loss function comparing to the proposed geometric only inner product, but its kernel lengthscales stay fixed throughout the optimization. In addition, KC discretizes the space to avoid the quadratic time cost, while our methods remain continuous with the help of GPU parallel computations.

\subsection{Deep Learning in Registration}
Fully connected layers with symmetric operations (max-pooling) in PointNet ~\cite{qi2017pointnet, aoki19pointnetlk}, convolutions of sparse tensors in FCGF\cite{FCGF2019}, sparse bilateral convolution in SPLATNet\cite{su18splatnet}, and graph convolution layers in DGCNN\cite{dgcnn19} can capture local and global geometric features of point clouds. Examples of utilizing deep geometric features include PCRNet\cite{Sarode2019PCRNetPC}, 3D-Feat-Net\cite{yew20183dfeat}, PointNetLK\cite{aoki19pointnetlk}.  

Given extracted features, point correspondences are calculated in the many-to-many~\cite{elbaz20173d,wang2019dcp} or one-to-one way\cite{jang2016categorical,paris20_3dregnet}. Correspondences of a point can be interpreted as a probabilistic distribution of its nearby points, predicted by convolutions and softmax operations over those points' feature embeddings~\cite{lu19deepicp}.  DCP~\cite{wang2019dcp} directly multiplies two feature embedding vectors between  all point pairs of the two point clouds, followed by softmax operations to get the correspondences. Deep Global Registration~\cite{choy2020deep} adopts convolution layers that take a candidate pair of points $(x,z)=(x_1, x_2, x_3, z_1, z_2, z_3)$ as input, and classifies whether this pair of point lies in a lower-dimensional manifold.

This work is not an end-to-end deep learning solution, but our unified point cloud function representation can incorporate deep learning features, such as semantics, into the cost function. The potential way of using our inner product as a loss of an  end-to-end framework can be a future study.

\section{Problem Setup}
\label{sec:problem}

Consider two (finite) collections of points, $X=\{x_i\}$, $Z=\{z_j\}\subset \mathbb{R}^3$. We want to determine which element \mbox{$h\in \mathrm{SE}(3)$}, aligns the two point clouds $X$ and $hZ = \{hz_j\}$ the ``best.'' To assist with this, we will assume that each point contains information described by a point in an inner product space, $(\mathcal{I},\langle\cdot,\cdot\rangle_{\mathcal{I}})$. To this end, we will introduce two labeling functions, $\ell_X:X\to\Ical$ and $\ell_Z:Z\to\Ical$.

 To measure their alignment, we turn the clouds, $X$ and $Z$, into functions $f_X,f_Z:\mathbb{R}^3\to\Ical$ that live in some reproducing kernel Hilbert space, $(\Hcal,\langle\cdot,\cdot\rangle_{\mathcal{H}})$. The action, $\mathrm{SE}(3) \curvearrowright \mathbb{R}^3$ induces an action $\mathrm{SE}(3) \curvearrowright \Hcal$ by $h.f(x) := f(h^{-1}x)$. Inspired by this observation, we will set $h.f_Z := f_{h^{-1}Z}$.

\begin{problem}\label{prob:problem}
	The problem of aligning the point clouds can now be rephrased as maximizing the scalar products of $f_X$ and $h.f_Z$, i.e., we want to solve
	\begin{equation}\label{eq:max}
		\argmax_{h\in \mathrm{SE}(3)} \, F(h),\quad F(h):= \langle f_X, f_{h^{-1}Z}\rangle_{\mathcal{H}}.
	\end{equation}
\end{problem}
\subsection{Constructing The Functions}

We follow the same steps in~\cite{MGhaffari-RSS-19} with an additional step in which we use the kernel trick to kernelize the information inner product. For the kernel of our RKHS, $\Hcal$, we first choose the squared exponential kernel $k:\mathbb{R}^3\times\mathbb{R}^3\to\mathbb{R}$:
\begin{equation}\label{eq:k}
k(x,z) = \sigma^2\exp\left(\frac{-\lVert x-z\rVert_3^2}{2\ell^2}\right),
\end{equation}
for some fixed real parameters (hyperparameters) $\sigma$ and $\ell$ (the \textit{lengthscale}), and $\lVert\cdot\rVert_3$ is the standard Euclidean norm on $\mathbb{R}^3$. This allows us to turn the point clouds to functions via
\begin{align}
	\nonumber f_X(\cdot) &:= \sum_{x_i\in X} \, \ell_X(x_i) k(\cdot,x_i), \\
	f_{h^{-1}Z}(\cdot) &:= \sum_{z_j\in Z} \, \ell_Z(z_j) k(\cdot,h^{-1}z_j).
\end{align}
Here $\ell_X(x_i)$ encodes the appearance information, for example LIDAR intensity and image pixel color. $k(\cdot, x_i)$ encodes the geometric information. We can now obtain the inner product of $f_X$ and $f_Z$ as
\begin{equation}\label{eq:scalar}
\langle f_X,f_{h^{-1}Z}\rangle_{\Hcal} := \sum_{\substack{x_i\in X, z_j\in Z}} \langle \ell_X(x_i),\ell_Z(z_j)\rangle_{\mathcal{I}} \cdot k(x_i,h^{-1}z_j)
\end{equation}

We use the kernel trick in machine learning\cite{bishop2006pattern,rasmussen2006gaussian,murphy2012machine} to substitute the inner products in~\eqref{eq:scalar} with the appearance  kernel.
After applying the kernel trick to~\eqref{eq:scalar}, we get
\begin{equation}
\label{eq:newscalar}
 \langle f_X,f_{h^{-1}Z}\rangle_{\Hcal} = \sum_{\substack{x_i\in X, z_j\in Z}}  k_c(\ell_X(x_i),\ell_Z(z_j)) \cdot k(x_i,h^{-1}z_j) ,
\end{equation}
We choose $k_c$ to be the squared exponential kernel with real hyperparameters $\sigma_c$ and $\ell_c$ that are set independently.

\begin{figure*}[t]
    \includegraphics[width=\columnwidth]{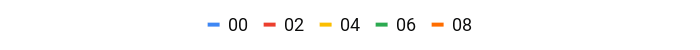}
    \hfill
    \includegraphics[width=\columnwidth]{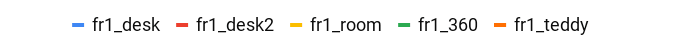}
    \includegraphics[width=0.5\columnwidth]{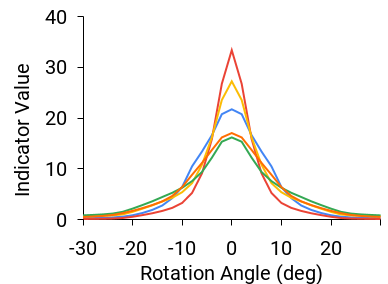}
    \includegraphics[width=0.5\columnwidth]{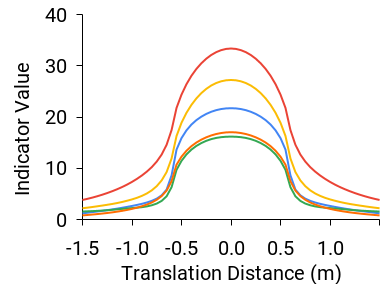}
    \includegraphics[width=0.5\columnwidth]{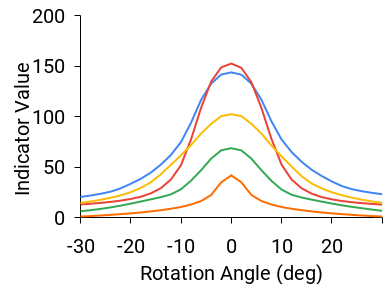}
    \includegraphics[width=0.5\columnwidth]{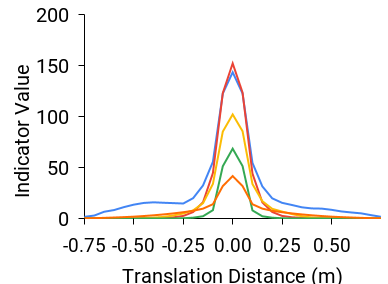}
    \caption{Indicator value with respect to rotation angle and translation distance for KITTI Stereo (left figures) and TUM RGB-D (right figures) sequences.} 
    \label{fig:stereo_indicator}
\end{figure*}

\subsection{Feature Embedding via Tensor Product Representation}
\label{sec:encode_feature_space}

We now extend the feature space to a hierarchical distributed representation. Let $(V_1, V_2, \dots)$ be different inner product spaces describing different types of non geometric features of a point, such as color, intensity, and semantics. Their tensor product, $V_1 \otimes V_2 \otimes \dots $ is also an inner product space. For any   $x\in X, z\in Z$ with features $\ell_X(x)=(u_1, u_2, \dots)$ and $\ell_Z(z)=(v_1, v_2, \dots)$, with $u_1, v_1 \in V_1$,  $u_2, v_2\in V_2$, $\dots$, we have
\begin{align}
    \label{eq:tensor_product}
    \nonumber \langle \ell_X(x), \ell_Z(z) \rangle_{\mathcal{I}} &= \langle u_1 \otimes u_2 \otimes \dots , v_1 \otimes v_2 \otimes \dots \rangle \\
    &= \langle u_1 , v_1 \rangle \cdot \langle u_2, v_2  \rangle  \cdot \dots .
\end{align}
By substituting~\eqref{eq:tensor_product} into \eqref{eq:scalar}, we obtain \begin{align}
    \nonumber \langle f_X,f_{h^{-1}Z}\rangle_{\Hcal} &= \sum_{\substack{x_i\in X\\z_j\in Z}} \langle u_{1i} , v_{1j} \rangle \cdot \langle u_{2i}, v_{2j}  \rangle   \dots  k(x_i, h^{-1}z_j) 
\end{align}
After applying the kernel trick we arrive at
\begin{align}
    \nonumber \langle f_X,f_{h^{-1}Z}\rangle_{\Hcal} &= \sum_{\substack{x_i\in X, z_j\in Z}} \, k(x_i,h^{-1}z_j) \cdot \prod_k k_{V_k} (u_{ki}, v_{kj}) \\
    &:= \sum_{\substack{x_i\in X, z_j\in Z}} \,k(x_i,h^{-1}z_j) \cdot c_{ij} . \label{eq:double_sum}
\end{align}

Equation~\eqref{eq:double_sum} describes the full geometric and non-geometric relationship between the two point clouds. Each $c_{ij}$ does not depend on the relative transformation, thus it will be a constant when computing the gradient and the step size. In our implementation, the double sum in ~\eqref{eq:double_sum} is sparse, because a point $x_i\in X$ is far away from the majority of the points $z_j \in Z$, either in the spatial (geometry) space or one of the feature (semantic) spaces.

This formulation can be further simplified to a purely geometric model, if we let the label functions $ \ell_X(x_i)= \ell_Z(z_j)=1$. Then  \eqref{eq:double_sum} becomes 
\begin{align}
\label{eq:geometric_only}
\langle f_X,f_{h^{-1}Z}\rangle_{\Hcal}=\sum_{\substack{x_i\in X, z_j\in Z}} \,k(x_i,h^{-1}z_j) .
\end{align}
Through~\eqref{eq:geometric_only}, the proposed method can register point clouds that do not have appearance measurements. It is worth noting that, when choosing the squared exponential kernel, \eqref{eq:geometric_only} has the same formulation as Kernel Correlation~\cite{tsin2004correlation}.

\subsection{An Indicator of Alignment}
\label{subsec:indicator}
We want to have an indicator  that represents the alignment of two point clouds $X$ and $Z$. An intrinsic metric available in our framework is the angle, $\theta$, between two functions. This indicator can be computed to track the optimization progress. The cosine of the angle is defined as
\begin{equation}
\label{eq:indicator}
 \cos(\theta) = \dfrac{\langle f_X,f_Z\rangle_{\Hcal}}{\lVert f_X \rVert \cdot \lVert f_Z \rVert} .
\end{equation}
However, calculating $\lVert f_X \rVert$ and $\lVert f_Z \rVert$ is time-consuming as it requires evaluating the double sum for each of the two point clouds. To approximate~\eqref{eq:indicator}, we use the following result. 
\begin{remark}
Suppose $k(x_i,x_j)=\delta_{ij}$ and $c_{ii}=1$, where $\delta_{ij}$ is the Kronecker delta, then $\lVert f_X\rVert = \sqrt{|X|}$.
\end{remark}
\begin{corollary}
Using the previous assumption, we define the following alignment indicator.
\begin{equation}
\label{eq:indicator_actual}
    i_{\theta} := \dfrac{1}{\sqrt{|X|\cdot |Z|}} \sum_{\substack{x_i\in X, z_j\in Z}}  c_{ij} \cdot k(x_i,z_j) .
\end{equation}
\end{corollary}
The behavior of the alignment indicator with respect to the rotation and translation errors is shown in Fig.~\ref{fig:stereo_indicator}. We manually rotate and translate the same point cloud and then calculate the indicator with the original point cloud. A larger transformation results in a smaller indicator value. Furthermore, the maximum indicator value occurs when the transformation error is zero. 

\begin{remark}
 OverlapNet~\cite{chen2020overlapnet} uses a neural network to predict a similar metric and detect loop closures. The cosine of the angle in~\eqref{eq:indicator} or the indicator in~\eqref{eq:indicator_actual} provide such a metric for \emph{self-supervised} learning while taking into account the semantic information. Given the promising results of ~\cite{chen2020overlapnet}, the combination of our metric with deep learning is an interesting future research direction. 
\end{remark}

\section{Considerations for Boosting the Performance}
\label{sec:practical_techniques}

The alignment indicator can guide the lengthscale update during the optimization. When the lengthscale is large, each point is associated with farther points, which provides the point cloud function representation a global perspective. When the lengthscale is small, each point is only connected to its closest neighbors, resulting in  local attention for the registration. For a single registration process, we use larger lengthscales at early iterations. Every time the alignment indicator value at the current lengthscale stabilizes, we decay the lengthscale by two percent. This mechanism is inspired by the deterministic annealing process in some GMM-based registration methods\cite{chui2000mpm,granger2002multiemicp,gold1998softassign}, which perform a fixed lengthscale decay every few iterations.

The hyperparameters to be tuned include the lengthscale of the geometric kernels and the appearance (color and semantic) kernels. We use the same hyperparameters within a dataset. At the first frame of an entire data sequence, we initialize the transformation to be identity. We use a large  lengthscale only for this single frame ($0.95$ in all the KITTI Stereo sequences), at the cost of more iterations. In subsequent frames, we use the previous estimated transformation as the initial value, accompanied by a smaller starting lengthscale ($0.1$ in all KITTI Stereo sequences).  

To address the costly double sum computation, we downsample the raw inputs. We adopt the FAST feature detector\cite{rosten2006FAST} implemented in OpenCV\cite{opencv_library}. We automatically control FAST's threshold of the pixel intensity difference and disable the non-max suppression, so that the number of selected pixels with non-empty depth values is between 3k and 15k.   

\section{Experimental Results}
\label{sec:exp}

We now present the frame to frame registration experiments on both outdoor and indoor datasets: KITTI stereo odometry \cite{behley2019semantickitti} and TUM RGB-D dataset \cite{sturm12iros}.

\subsection{Experimental Setup}
All experiments are performed in a frame-to-frame manner without skipping images. The first frame's transformation is initialized with identity, and all later frames start with the previous frames' results. Hyperparameters for the proposed method on KITTI stereo and TUM RGB-D are available at~\cite{unified-cvo}. The same values were used within one dataset.

\begin{table*}[t]
\begin{center}
\caption{Results of the proposed frame-to-frame method using the KITTI  \cite{kitti2012} stereo odometry
benchmark as evaluated on the average drift in translation, as a percentage (\%), and rotation, in degrees per meter($\degree/\m$). The drifts are calculated for all possible subsequences of $100,200....,800$ meters.
}
\scriptsize
\begin{tabular}{l l | r r r r r r r r r r r | r r}
\toprule
& & 00 & 01 & 02 & 03 & 04 & 05 & 06 & 07 & 08 & 09 & 10 &  \textbf{Avg} & \textbf{Std} \\
\midrule
GeometricCVO & t (\%) &  4.06 & 7.04 & 5.86 & 
\textbf{3.84} &  5.08 & 3.42 & \textbf{2.99} & 5.23 & 4.40 & 4.67 & \textbf{3.42} & 4.55 & 1.20 \\
&  r ($\degree/\m$) & 0.0173 &
0.0285 &
0.0220&
0.0199&
0.0358&
0.0206&
0.0151&
0.0444&
0.0188&
0.0185&
0.0181&
0.0236&
0.00907\\
\midrule
GICP \cite{segal2009gicp} &  t (\%) & 8.66 &
26.19 &
7.92 &
7.64 &
7.40 &
6.06 &
16.40 &
8.45 &
14.69 &
7.35 &
12.73 &11.23 &
5.99   \\
&  r ($\degree/\m$) & 0.0361 &
0.0467 &
0.0302 &
0.0460 &
0.0548 &
0.0336 &
0.0616 &
0.0657 &
0.0453 &
0.0248 &
0.0525 &
0.0452 &
0.0130  \\
\midrule
3D-NDT \cite{magnusson2007ndt3d} &  t (\%) & 7.53 &
16.41 &
6.11 &
5.13 &
4.63 &
6.76 &
11.68 &
11.16 &
7.67 &
5.50 &
10.96 &
8.50 &
3.63 \\
&  r ($\degree/\m$) & 0.0388 &
0.0272 &
0.0261 &
0.0432 &
0.0302 &
0.0346 &
0.0472 &
0.0791 &
0.0387 &
0.0237 &
0.0467 &
0.0396 &
0.0155\\
\midrule
\midrule
ColorCVO & t (\%) & \textbf{3.19} & 4.42 &
5.00 &
3.94 &
3.86 &
\textbf{2.94} &
3.18 &
\textbf{2.32} &
\textbf{3.65} &
4.39 &
3.64 &
3.69 &
0.76   \\
&  r ($\degree/\m$) & \textbf{0.0125} &
0.0158 &
0.0167 &
0.0182 &
0.0230 &
0.0152 &
\textbf{0.0103} &
0.0176 &
0.0147 &
0.0151&
0.0154 &
0.0159 &
0.00323 \\
\midrule
MC-ICP \cite{servos2014mcicp} &  t (\%) & 7.77 &
55.26 &
11.33 &
15.45 &
9.65 &
5.51 &
9.65 &
13.62 &
6.54 &
8.16 &
12.16 &
14.10 &
13.98  \\
&  r ($\degree/\m$) &
0.0387 &
0.0598 &
0.0357 &
0.0749 &
0.0585 &
0.0335 &
0.0335 &
0.0927 &
0.0314 &
0.0277 &
0.0504 &
0.0488 &
0.0208 \\
\midrule
\midrule
SemanticCVO & t (\%) & 3.22 &
\textbf{3.97} &
\textbf{4.96} &
3.94 &
\textbf{3.84} &
2.95 &
3.28 &
2.35 &
\textbf{3.65} &
\textbf{4.32} &
3.59 &
\textbf{3.64} &
\textbf{0.70}  \\
(with color)&  r ($\degree/\m$) & 0.0126 &
\textbf{0.0132} &
\textbf{0.0166} &
\textbf{0.0179} &
\textbf{0.0227} &
\textbf{0.0150} &
0.0105 &
\textbf{0.0172} &
\textbf{0.0146} &
\textbf{0.0148} &
\textbf{0.0151} &
\textbf{0.0155} &
\textbf{0.00321}\\

\bottomrule
\end{tabular}
\label{tab:kitti_stereo_reults}
\end{center}
\end{table*}

On KITTI, our baselines are  GICP~\cite{segal2009gicp}, Multichannel-ICP~\cite{servos2014mcicp}, and 3D-NDT~\cite{magnusson2007ndt3d}. GICP and NDT are compared with our geometric method (\textit{Geometric CVO}). Multichannel-ICP competes with our color-assisted method (\textit{Color CVO}). GICP and 3D-NDT implementation are from PCL~\cite{pcl}. The Multichannel-ICP implementation is from   \cite{parkison2019rkhsicp}. Both the baselines and the proposed methods remove the first 100 rows of image pixels that mainly include sky pixels, as well as points that are more than 55 meters away. All the baselines use full point clouds without downsampling. The discussions of more candidate baselines and point selectors are in Sec.~\ref{sec:discussion}

On TUM RGB-D, we use the same baselines for geometric registration as KITTI. We compare Color CVO with Dense Visual Odometry (DVO)~\cite{kerl2013dense} and Color ICP~\cite{park2017colored}. We reproduced DVO results with the code from \cite{MatthieuP2019DVO} because the original DVO source code requires an outdated ROS dependency~\cite{Kerl2013DVOrepo}. The Color ICP implementation is taken from Open3D~\cite{Zhou2018}. The baselines use full point clouds.

\begin{figure}[t]
  \centering 
  \subfloat{\includegraphics[width=0.99\columnwidth,clip]{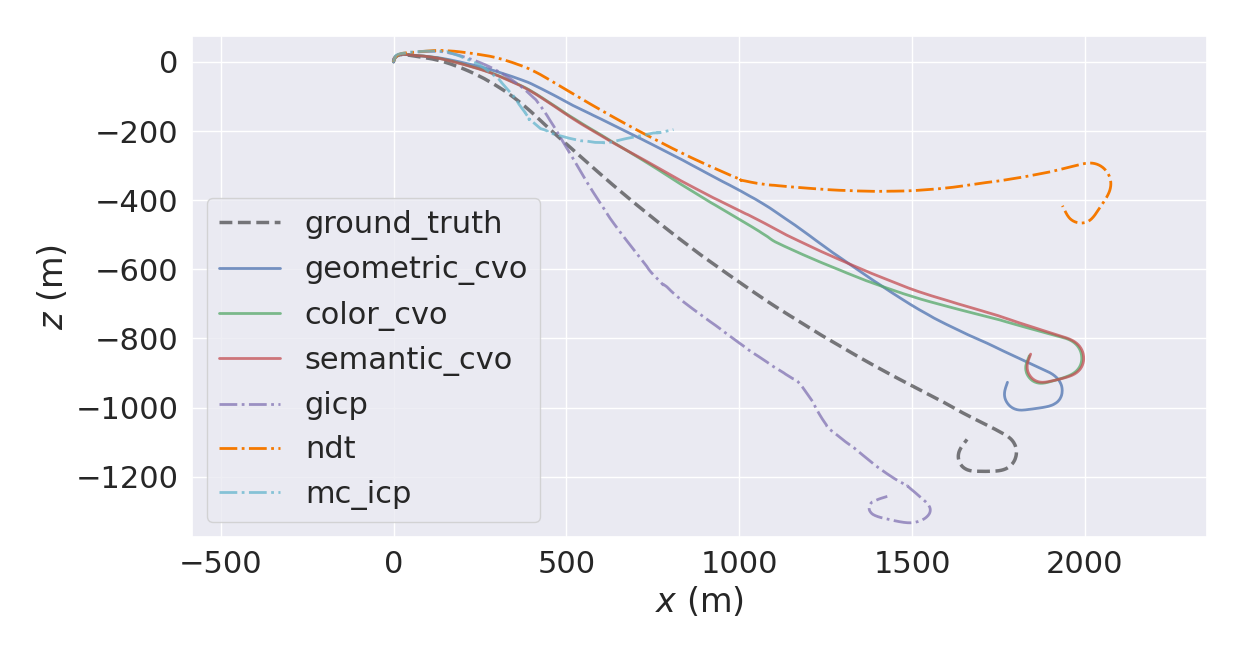}}\\
  \subfloat{\includegraphics[width=0.99\columnwidth,clip]{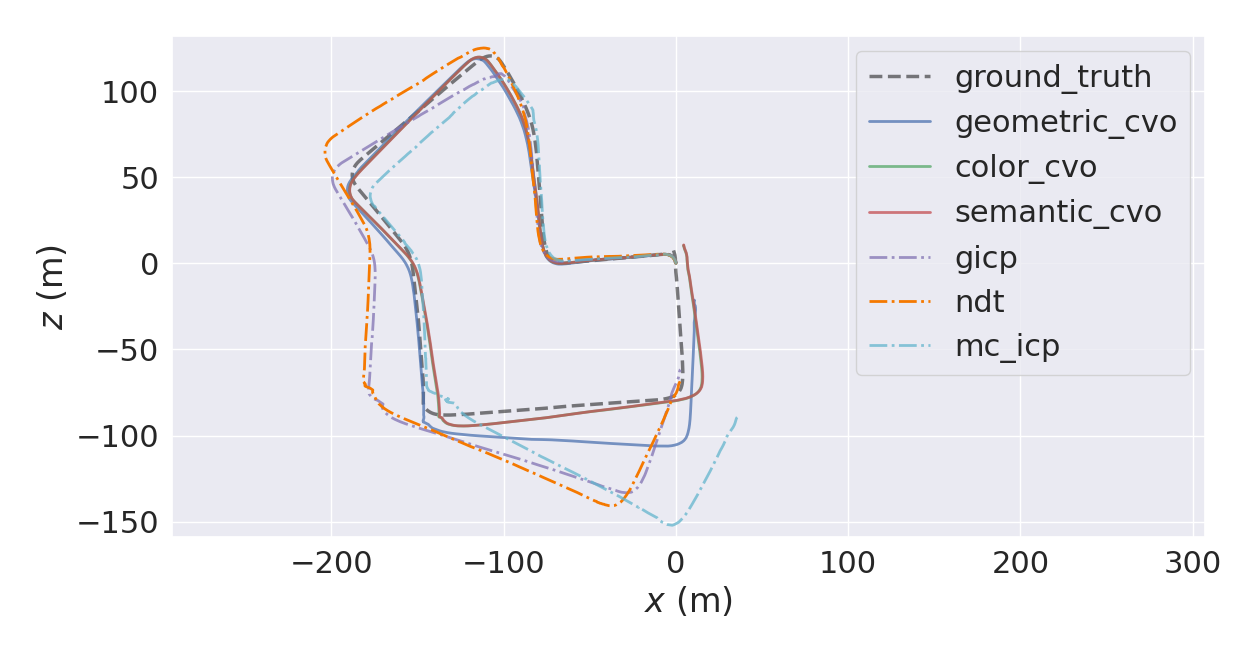}} 
  
  \caption{An illustration of the proposed registration method on KITTI stereo sequence 01 (top) and 07 (bottom) versus the baselines. The black dashed trajectory is the ground truth. The dot-dashed trajectories are the baselines. Plotted with EVO\cite{grupp2017evo}. }
 \label{fig:kitti01_07_stereo}
\end{figure}

\subsection{Outdoor Stereo Camera: KITTI Stereo Odometry}

We select a subset of pixels from KITTI's stereo images via OpenCV's FAST~\cite{rosten2006FAST} feature detector. The depth values of the selected pixels are generated with ELAS~\cite{libelas2010}. The semantic predictions of the images come from  Nvidia's pre-trained neural network~\cite{zhu2019improving}, which was trained on 200 labeled images. Examples of the point clouds are in Fig.~\ref{fig:difficult_scenes}.
Noise from the estimated depth, from the color sensor, and from the semantic predictions are visible. 

\begin{figure}[t]
  \centering 
  \includegraphics[width=0.99\columnwidth,trim={0.0cm 1.0cm 0.5cm 0.5cm},clip]{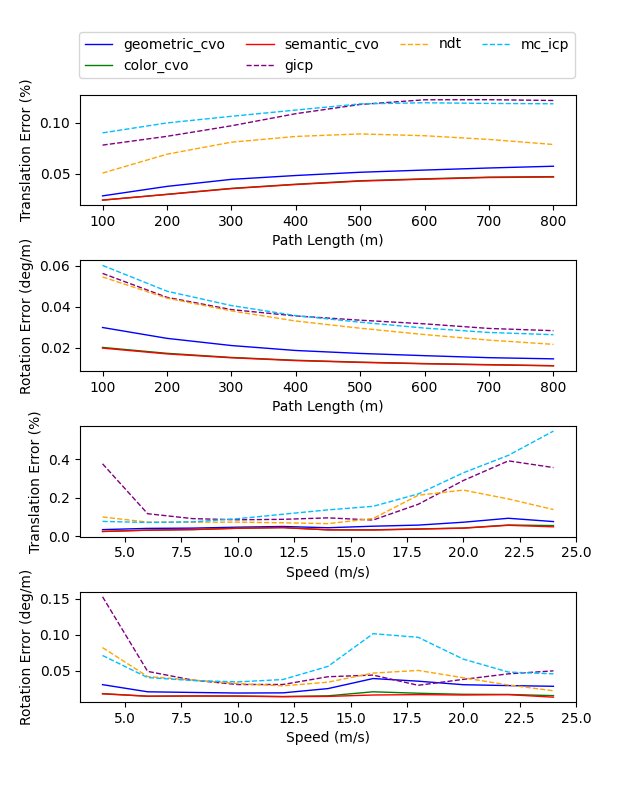} 
  \caption{From top to down: the average translation errors and rotation errors on KITTI Stereo sequences 00 to 10 with respect to the distance segment and the moving speed, respectively.}
 \label{fig:kitti_speed}
\end{figure}

The result of Geometric, Color, and Semantic CVO and other baselines are provided in Table \ref{tab:kitti_stereo_reults}. From sequence 00 to 10, our geometric  method  has a lower average  translational error ($4.55\%$) comparing to the GICP ($11.23\%$) and NDT ($8.50\%$). Our color version has a lower average translational drift ($3.69\%$) than Multichannel-ICP ($14.10\%$). If we add  semantic information the error is further reduced ($3.64\%$). The addition of color and semantic information also yields a lower standard deviation. Meanwhile, the average rotational drift of the proposed methods are smaller. Specifically, on the highway sequence (01) where the point cloud pattern becomes repetitive and noisy , both NDT and GICP perform poorly (as shown in Fig.~\ref{fig:kitti01_07_stereo}). Figure~\ref{fig:kitti_speed} shows the average translational and rotational errors at different distances and speeds. The proposed methods show a more consistent high accuracy across different speeds.

On our desktop computer, excluding the image I/O and point cloud generation operations, the current GPU implementation takes an average $1.4 \sec$ per frame on registering less than 15k points. GICP, NDT, Multichannel-ICP all use full point clouds (150k-350k points), and take $6.3 \sec$, $6.6 \sec$, and $57 \sec$ per frame, respectively.

\begin{table*}[t]
    \centering
    \caption{The RMSE of Relative Pose Error (RPE) for \texttt{fr1} sequences. The trans. columns show the RMSE of the translational drift in $\mathrm{m}/\sec$ and the rot. columns show the RMSE of the rotational error in $\mathrm{deg}/\sec$.}
    \scriptsize
    \begin{tabular}{lcc|cc|cc||cc|cc|cc}
        \toprule
          & \multicolumn{2}{c}{Geometric CVO} & \multicolumn{2}{c}{GICP\cite{segal2009gicp}} & \multicolumn{2}{c}{3D-NDT\cite{magnusson2007ndt3d}} & \multicolumn{2}{c}{Color CVO} & \multicolumn{2}{c}{DVO\cite{kerl2013dense}} & \multicolumn{2}{c}{Color ICP\cite{park2017colored}} \\
         Sequence & Trans. & Rot. &  Trans. & Rot. & Trans. & Rot. & Trans. & Rot. & Trans. & Rot. &  Trans. & Rot.\\
        \midrule
        fr1/desk    & 0.0493 & 2.3377 & 0.2358 & 11.9360 & 0.2404 & 13.5183 & 0.0384 & \bf2.1422 &
                     0.0387 & 2.3589 & 0.0938 & 5.2660 \\
        fr1/desk2   & 0.0545 & \bf2.7190 & 0.3617 & 19.8483 & 0.1823 & 11.8914 & \bf0.0515 & 2.8967 &
                     0.0518 & 3.6529 & 0.2304 & 8.5799 \\ 
        fr1/room    & 0.0565 & \bf2.2946 & 0.3966 & 17.0337 & 0.1718 & 9.9076 & \bf0.0501 & 2.3366 &
                     0.0518 & 2.8686 & 0.1444 & 6.2150 \\
        fr1/360     & \bf0.1001 & \bf2.8686 & 0.5251 & 17.0537 & 0.2245 & 13.6262 & 0.1021 & 3.1086 & 
                     0.1602 & 4.4407 & 0.2325 & 8.6135 \\
        fr1/teddy   & 0.0663 & \bf2.4122 & 0.4659 & 16.3678 & 0.2095 & 11.2214 & 0.0668 & 2.6016 &
                     0.0948 & 2.5495 & 0.1735 & 5.7976 \\ 
        fr1/floor   & 0.2267 & 2.7345  & 0.2008 & 6.5601 & 0.5560 & 35.9573 & 0.0697 & 
                    2.3663 &
                     \bf0.0635 & \bf2.2805 & 0.0668 & 3.3416 \\ 
        fr1/xyz     & \bf0.0238 & \bf0.9748 & 0.1093 & 7.8490 & 0.1102 & 5.5953 & 0.0270 & 1.1379 &
                     0.0327 & 1.8751 & 0.0632 & 4.5334 \\
        fr1/rpy     & 0.0413 & 3.1806  & 0.4802 & 19.4342 & 0.2329 & 16.8113 & 0.0501 & 3.6598 &
                     0.0336 & \bf2.6701 & 0.0930 & 5.8095 \\
        fr1/plant   & 0.0388 & 1.9027 & 0.8551 & 26.8711 & 0.1335 & 7.7507 & 0.0347 & 1.6451 &
                     \bf0.0272 & \bf1.5523 & 0.1205 & 4.9295 \\
        \midrule
        Average & 0.0730 & \bf2.3805 & 0.4034 & 15.8838 & 0.2290 & 14.0311 & \bf0.0545 & 2.4333 &
                     0.0623 & 2.6943 & 0.1353 & 5.8985 \\
        \bottomrule
    \end{tabular}
    \label{tab:fr1_results}
\end{table*}

\begin{table*}[t]
    \centering
    \caption{The RMSE of Relative Pose Error (RPE) for the structure v.s texture sequence. The Trans. columns show the RMSE of the translational drift in $\mathrm{m}/\sec$ and the Rot. columns show the RMSE of the rotational error in $\mathrm{deg}/\sec$.}
    \scriptsize
    \begin{tabular}{lllcc|cc|cc||cc|cc|cc}
        \toprule
          & & & \multicolumn{2}{c}{Geometric CVO} & \multicolumn{2}{c}{GICP\cite{segal2009gicp}} & \multicolumn{2}{c}{3D-NDT\cite{magnusson2007ndt3d}} & \multicolumn{2}{c}{Color CVO} & \multicolumn{2}{c}{DVO\cite{kerl2013dense}} & \multicolumn{2}{c}{Color ICP\cite{park2017colored}} \\
         \multicolumn{3}{l}{structure-texture} & Trans. & Rot. & Trans. & Rot. & Trans. & Rot. & Trans. & Rot.  & Trans. & Rot.& Trans. & Rot. \\
        \midrule
        $\times$   & \checkmark & near & 0.0267 & 0.8745 & 0.2602 & 7.5238 & 0.4586 & 13.4089 & 0.0250 & \bf0.8201 &
                                         0.0563 & 1.7560 & \bf0.0212 & 0.9744\\
        $\times$   & \checkmark & far  & \bf0.0498 & 1.1602 & 0.3115 & 3.3421 & 0.2034 & 4.8534  & 0.0591 & \bf1.1393 & 
                                        0.1612  & 3.4135 & 0.0755 & 1.6356\\
        \checkmark & $\times$   & near & 0.0338 & 2.4081 & 0.0628 & 2.0061 & 0.0993 & 5.5899 & 0.0505 & 3.5577 & 
                                        0.1906  & 10.6424 & 0.0255 & \bf1.0317\\
        \checkmark & $\times$   & far  & \bf0.0376 & 1.2435 & 0.1172 & 3.6457 & 0.0861 & 1.8595 &  0.0456 & \bf1.2239 & 
                                        0.1171  & 2.4044 & 0.0592 & 1.7822\\
        \checkmark & \checkmark & near & 0.0238 & 1.3058 & 0.1573 & 6.0924 & 0.1082 & 4.6971 & 0.0344 & 1.6899 & 
                                        \bf0.0175  & \bf0.9315 & 0.0200 & 1.2008 \\
        \checkmark & \checkmark & far  & 0.0288 & 0.9314 & 0.1921 & 4.6908 & 0.0717 & 1.9343 &  0.0293 & 0.9516 & 
                                        \bf0.0171  & \bf0.5717 & 0.0434 & 1.1375\\
        $\times$   & $\times$   & near & 0.3057 & 10.8878 & 0.3685 & 12.6208 & 0.5901 & 16.1501 &  0.2143 & 8.9564 &
                                        0.3506  & 13.3127 & \bf0.2064 & \bf7.7856 \\
        $\times$   & $\times$   & far  & \bf0.1287 & 4.0173 & 0.2232 & 2.4611 & 0.3722 & 7.3946  &   0.1449 & 2.9821 & 
                                        0.1983  & 6.8419 & 0.2052 & \bf2.0850\\
        \midrule
		\multicolumn{3}{c}{Average} & 0.0794 & 2.8536 & 0.2116 & 5.2979 & 0.2487 & 6.9860 &  \bf0.0754 & 2.6651 & 
		                                  0.1386  & 4.9843 & 0.0820 & \bf2.2041\\
        \bottomrule
    \end{tabular}
    \label{tab:fr3_results}
\end{table*}

\subsection{Indoor RGB-D Camera: TUM RGB-D Dataset}
For TUM RGB-D, a semi-dense point cloud is generated from the depth images with FAST~\cite{rosten2006FAST} feature selector. We evaluated our method on the \texttt{fr1} sequences, which are recorded in an office environment, and \texttt{fr3} sequences, which contain image sequences in structured/nostructured and texture/notextured environments. TABLE~\ref{tab:fr1_results} shows the results of \texttt{fr1} sequences. Geometric CVO outperforms the baselines and achieves a similar performance to DVO. Moreover, with color information, the average error of CVO decreases.

The results of \texttt{fr3} sequences is shown in TABLE~\ref{tab:fr3_results}. CVO outperforms the baselines. The overall result of Color CVO is better than Geometric CVO. However, Geometric CVO has lower translation errors in some sequence. This might because of the motion blur effect in the image, where color information is noisy due to rapid camera motion. 

\section{Discussions and Limitations}
\label{sec:discussion}
Besides the reported baseline results, we tried to run Semantic ICP~\cite{parkison2018semanticicp} and Color ICP\cite{park2017colored} with KITTI's full stereo point clouds as well. However, Semantic-ICP takes $4-8\min$ per frame on our machine, thus is infeasible to complete all the 23190 frames. The original Color ICP work has not been tuned for stereo data, and  failed on KITTI sequence 00 and 08. We also tried to use the FAST point selector  for all the baselines, but only GICP shows  improvements, with $7.98\%$ translation drift and $0.0362\degree/\m$ rotation drift, versus our geometric result being $4.55\%$ and $0.0236\degree/\m$.

We noticed that the point selector has a significant influence on the performance of the proposed methods. DSO's semi-dense point selector in~\cite{engel2016dso} was unable to complete some challenging sequences such as KITTI sequence 01. We cannot use PCL's Voxel Filter\cite{pcl} either because the original color and semantic information is lost during its downsampling. Only FAST\cite{rosten2006FAST} feature selector from OpenCV\cite{opencv_library} works for all the datasets we tested. A future direction is to find a more robust  downsampling scheme for this framework.

Moreover, the proposed methods' performance relies on the geometric lengthscale during the optimization. Adaptive CVO\cite{lin2019adaptive} addresses the lengthscale decay by regarding it as a part of the optimizing variable. Still we need to manually choose an initial lengthscale. For inputs with larger accelerations, the lengthscale needs a global perspective for such  abrupt changes. In this case, another future direction is an algorithmic way of selecting the initial lengthscale, or more broadly, studying the hyperparameter learning problem. 

\section{Conclusion}
\label{sec:conclusion}
We developed a nonparametric registration framework that integrates geometric and semantic measurements and does not require explicit data association. The proposed approach can utilize the extra visual and semantic information from modern range sensors while not being restricted by pairwise data correspondences. The novel hierarchical distributed representation of features via the tensor product representation provides a mathematically sound and systematic way of incorporating semantic knowledge into the point cloud model. The evaluations using publicly available stereo and RGB-D datasets show that the proposed method outperforms state-of-the-art outdoor and indoor frame-to-frame registration methods. We also provided an open-source GPU implementation. 
 
In the future, we shall explore the connections of the developed framework with deep learning for representation learning in applications such as multi-modal feature learning, place recognition, and robust tracking and SLAM in harsh and visually degraded situations.

\bibliographystyle{IEEEtran}
\bibliography{strings-abrv,ieee-abrv,refs}

\end{document}